\documentclass[conference]{IEEEtran}
\IEEEoverridecommandlockouts
\input epsf
\usepackage{graphicx}
\usepackage{cite}
\usepackage{amssymb}
\usepackage{amsmath}
\usepackage{float}
\usepackage{booktabs}
\usepackage{tikz}
\usetikzlibrary{shapes,positioning}
\usepackage{multirow}

\def\BibTeX{{\rm B\kern-.05em{\sc i\kern-.025em b}\kern-.08em
    T\kern-.1667em\lower.7ex\hbox{E}\kern-.125emX}}

\begin{document}

\makeatletter
\newcommand{\newlineauthors}{%
  \end{@IEEEauthorhalign}\hfill\mbox{}\par
  \mbox{}\hfill\begin{@IEEEauthorhalign}
}
\makeatother

\title{Credit Card Fraud Detection with Subspace Learning-based One-Class Classification\\}

\author{\IEEEauthorblockN{Zaffar Zaffar\IEEEauthorrefmark{1},
Fahad Sohrab\IEEEauthorrefmark{1},
Juho Kanniainen\IEEEauthorrefmark{1} 
and
Moncef Gabbouj\IEEEauthorrefmark{1}}
\IEEEauthorblockA{\IEEEauthorrefmark{1}Faculty of Information Technology and Communication Sciences, Tampere University, Finland}

{\{zaffar.zaffar, fahad.sohrab, juho.kanniainen, moncef.gabbouj\}}@tuni.fi}

\maketitle

\begin{abstract}
In an increasingly digitalized commerce landscape, the proliferation of credit card fraud and the evolution of sophisticated fraudulent techniques have led to substantial financial losses. Automating credit card fraud detection is a viable way to accelerate detection, reducing response times and minimizing potential financial losses. However, addressing this challenge is complicated by the highly imbalanced nature of the datasets, where genuine transactions vastly outnumber fraudulent ones. Furthermore, the high number of dimensions within the feature set gives rise to the ``curse of dimensionality". In this paper, we investigate subspace learning-based approaches centered on One-Class Classification (OCC) algorithms, which excel in handling imbalanced data distributions and possess the capability to anticipate and counter the transactions carried out by yet-to-be-invented fraud techniques. The study highlights the potential of subspace learning-based OCC algorithms by investigating the limitations of current fraud detection strategies and the specific challenges of credit card fraud detection. These algorithms integrate subspace learning into the data description; hence, the models transform the data into a lower-dimensional subspace optimized for OCC. Through rigorous experimentation and analysis, the study validated that the proposed approach helps tackle the curse of dimensionality and the imbalanced nature of credit card data for automatic fraud detection to mitigate financial losses caused by fraudulent activities.
\end{abstract}

\IEEEoverridecommandlockouts
\IEEEpeerreviewmaketitle

 \section{Introduction} \label{sec:introduction}
The Federal Trade Commission has recently reported an alarming increase in credit card fraud reports and the revenue lost due to such frauds in the past few years \cite{Public_tableau}. One of the key factors of such an increase in credit card fraud is the digitalization of commerce because of the COVID-19 outbreak and the shutdown of the whole world \cite{habibpour2023uncertainty,10.1145/3474379}. Credit card fraud has existed since the invention of payment cards, and different policies were formulated and brought into practice from time to time to reduce the losses incurred by such frauds. The address verification system, keeping a scoring record of positive and negative lists to identify and prevent high-risk transactions \cite{quah2008real}, and the use of Card Verification Value (CVV) by Visa and Card Verification Code (CVC) by MasterCard \cite{barker2008credit} are a few of the examples of the preventive policies. As for the detective approach, many Machine Learning (ML) models, such as Support Vector Machines (SVM), logistic regression, random forest \cite{BHATTACHARYYA2011602}, artificial neural networks, k-nearest neighbors (kNN) \cite{asha2021credit} and Self-Organizing Maps (SOM) \cite{zaslavsky2006credit}, have been implemented for this cause. The uptrend in fraud cases and lost revenue, despite these policies, clearly shows that the previous set of measures, both preventive and detective, is not enough. 

In order to have a better solution that can effectively and efficiently mitigate the losses due to these frauds, we have to understand the shortcomings of the previous approaches as well as the challenges in the credit card fraud detection problem in general. Credit card fraud detection is a binary classification problem having two classes: normal (or positive class) and fraudulent (or negative class). A very basic property and one of the main issues in these problems is that the data is highly imbalanced \cite{dal2014learned}, owing to the fact that billions of card transactions take place every month worldwide, and a significantly smaller amount of transactions are fraudulent. To deal with the data imbalance issue, the ML models that are in practice have used sampling techniques; that is, a sample from the majority class, based on some sampling criterion, is taken \cite{zhang2019gmm} or instances for the minority class are synthetically generated based on some criterion \cite{sisodia2017performance} so that the number of instances in both classes is made equal. In some cases, an approach based on both of the sampling techniques is used to have a balanced dataset \cite{ahammad2020credit}. 

Another property of fraud detection problems is that the fraudulent activities and techniques evolve with time \cite{kulatilleke2022challenges}. Any method used by the fraudsters is identified by the anti-fraud team of the respective organization, and efforts are made to stop further losses through the same fraudulent technique. Consequently, personnel with the aim of gaining unlawful advantage of people or systems (or both) try to come up with new ideas and techniques. The ML algorithms that have been implemented for this purpose can only model the fraudulent techniques that are already in practice; that is, they cannot model, and hence, detect, the fraud that will be carried out by methods that are not existent and are yet to be invented. Therefore, we need a model that can also detect, predict, and stop fraud by the methods that will be invented in the future. 

One-Class Classification (OCC) algorithms, on the other hand, take data from only a single (positive or the normal) class for training, which is usually available in abundance, and form a boundary around the positive class (or between the two classes). These algorithms classify everything that lies outside the inferred boundary as a negative class object. These algorithms have been implemented in many different domains and have proved to be a good solution with good performance for the respective problem. The examples of such domains include but are not limited to bot detection on Twitter \cite{RODRIGUEZRUIZ2020101715}, spoofing detection \cite{6712706}, \cite{ying2023gps}, video surveillance \cite{10.1007/978-3-642-15552-9_28}, machine fault detection for predictive maintenance \cite{shin2005one}, hyper-spectral image analysis and classification \cite{kilickaya2023hyperspectral}, and Myocardial Infarction (MI) detection \cite{10081907}.

To address the above-mentioned challenges in fraud detection problems and to resolve the curse of dimensionality by embedding the feature extraction into the training phase of the algorithm and letting the model extract a discriminative set of features, we propose to use a set of OCC algorithms that are ideal for the highly imbalanced dataset and can effectively model and detect the fraudulent transactions carried out by to-be-invented techniques. For this purpose, we experimented with several OCC models to find a more efficient way to reduce the losses by such frauds.

\begin{figure*}[htp]
    \centering
    \includegraphics[width=14cm]{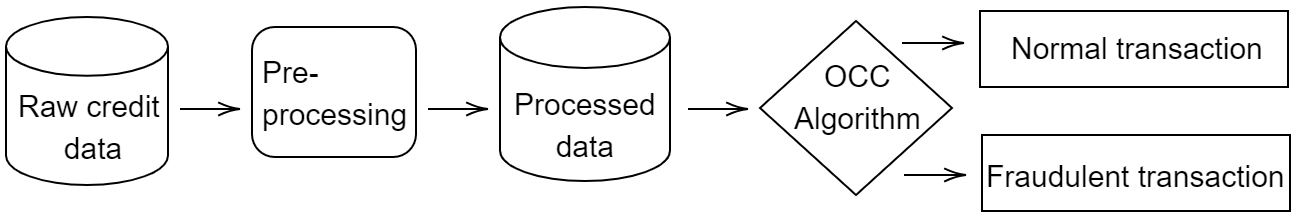}
    \caption{The flowchart depicting the credit card detection system.}
    \label{fig:flowchart}
\end{figure*}

\section{Methodology} \label{sec:methodology}
In the OCC setting, data from the target (positive) class is used to develop an optimal boundary between the target data and outliers. Depending on the model, the structure of the decision boundary varies. For instance, the One-Class Support Vector Machine (OCSVM) has a hyperplane \cite{senf2006comparison}, the Support Vector Data Description (SVDD) has a hyper-sphere \cite{Tax2004}, and the Ellipsoidal Subspace Support Vector Data Description (ESSVDD) has an ellipsoidal boundary \cite{9133428} differentiating the two class (fraudulent and normal transaction) data from each other. A general overview of the credit card fraud detection system with an OCC algorithm is depicted in Figure \ref{fig:flowchart}. The Subspace Support Vector Data Description (SSVDD), is the SVDD-based model where the data is projected, using a projection matrix $\mathbf{Q}$, from the original \textit{D} dimensions to the lower \textit{d}-dimensional subspace iteratively during the training \cite{sohrab2018subspace}. $\mathbf{Q}$, incorporated with the matrix $\mathbf{S_{Q}}$, representing the geometric information of the data in the subspace, is employed to find the optimized set of features in the Graph-embedded Subspace Support Vector Data Description (GESSVDD) \cite{sohrab2023graph}. For data vectors represented by $\mathbf{x}_{i} \in \mathbb{R}^{D}$, where $i = 1,2,...,N$, the mathematical formulation of the GESSVDD problem is as follows:

\begin{equation}
    \begin{split}
        \label{gessvdd_prob}
        & min \quad R^{2} + C \sum_{i=1}^{N} \xi_{i} \\
        & s.t.: \left \| \mathbf{S_{Q}}^{-\frac{1}{2}}\mathbf{Qx}_{i} - \mathbf{u} \right \|_{2}^{2} \leq R^{2} + \xi_{i}, \\
        & \xi_{i} \geq 0 \quad i=1,2,...,N,
    \end{split}
\end{equation}
where \textit{N} is the number of data points, $R$ is the radius, and \(\mathbf{u=S_{Q}}^{-\frac{1}{2}}\mathbf{a}\) is the center of the hyper-sphere in the subspace ($\mathbf{a}$ is the center in the original feature space). The variable $\xi_{i}$ represents the slack variables, and \textit{C} denotes the trade-off between maximizing the margin (enclosing more data points in the boundary) and minimizing the radius. To solve the optimization problem in \eqref{gessvdd_prob}, it is reformulated into a Lagrangian function using the Lagrange multipliers $\alpha_{i}$ and $\gamma_{i}$.

\begin{align}\label{gessvdd_lag}
        L &=  R^{2} + C\sum_{i=1}^{N}\xi_{i} - \sum_{i=1}^{N}\alpha_{i} \left( \right. R^{2} + \xi_{i} \notag\\
        &- (\mathbf{S_{Q}}^{-\frac{1}{2}}\mathbf{Qx}_{i})^{T}\mathbf{S_{Q}}^{-\frac{1}{2}}\mathbf{Qx}_{i} + 2\mathbf{u}^{T}\mathbf{S_{Q}}^{-\frac{1}{2}}\mathbf{Qx}_{i} - \mathbf{u}^{T}\mathbf{u} \left. \right ) \notag\\ 
        &- \sum_{i=1}^{N}\gamma_{i}\xi_{i}.
\end{align}

The solution of \eqref{gessvdd_lag} provides us with the $\alpha_{i}$ values for each instance in the dataset. These $\alpha_{i}$ values, representing the position of a data point in the projected subspace, are important for determining $\mathbf{u}$ and $R$ of the hyper-sphere. If an $\alpha$ value is zero, the data point lies inside the boundary of the hyper-sphere. If an $\alpha$ value falls between 0 and the regularization parameter \textit{C}, such data point, denoted by $\mathbf{s}$, lies on the boundary of the hyper-sphere and is known as a support vector. On the other hand, if an $\alpha$ value exceeds \textit{C}, the data point lies outside the boundary of the hyper-sphere. The radius of the optimal hyper-sphere can be calculated using

\begin{equation}
    \label{gessvdd_rad}
    R = \sqrt{(\mathbf{S_{Q}}^{-\frac{1}{2}}\mathbf{Qs})^{T} \mathbf{S_{Q}}^{-\frac{1}{2}}\mathbf{Qs} - 2(\mathbf{S_{Q}}^{-\frac{1}{2}}\mathbf{Qs})^{T}\mathbf{u} + \mathbf{u}^{T}\mathbf{u}}.
\end{equation}

To classify any test data vector $\mathbf{x}^{*}$ into its respective class, it must first be transformed to the lower \textit{d}-dimensional subspace using the same $\mathbf{Q}$ and $\mathbf{S_{Q}}$ and the distance of $\mathbf{x}^{*}$ from $\mathbf{u}$ in transformed feature subspace is calculated and checked if it is greater or smaller than the $R$ given in \eqref{gessvdd_rad}. It is classified as a non-fraudulent transaction if it satisfies the following decision rule:

\begin{equation}
    \label{gessvdd_test}
    \left \| \mathbf{S_{Q}}^{-\frac{1}{2}}\mathbf{Qx}^{*} - \mathbf{u} \right \|_{2}^{2} \leq R^{2}.
\end{equation}

The matrix $\mathbf{S_{Q}}$, having geometric information in the data in the transformed feature subspace, is mathematically represented as:

\begin{equation}
    \label{gessvdd_s}
    \mathbf{S_{Q}} = \mathbf{QXL}_{x}\mathbf{X}^{T}\mathbf{Q}^{T} = \mathbf{QS}_{x}\mathbf{Q}^{T},
\end{equation}
where $\mathbf{X} \in \mathbb{R}^{N \times D}$ is the data matrix and $\mathbf{L}$ is the matrix representation of the graph. Based on the choice of the $\mathbf{L}$ in \eqref{gessvdd_s}, there can be many variants of the model. In this study, we have implemented three GESSVDD variants by considering different options for $\mathbf{L}$. These are:

\begin{itemize}
    \item The first variant, denoted as GESSVDD-I, replaces $\mathbf{L}_{x}$ with the identity matrix, $\mathbf{I}$. 
    
    \item The second variant, referred to as GESSVDD-PCA, utilizes the Principal Component Analysis (PCA) graph where $\mathbf{S}_{x}$ is replaced with \( \frac{1}{N}\mathbf{S}_{t}\). The Scatter matrix, $\mathbf{S}_{t}$ is derived as
    \begin{equation}
        \label{gessvdd_totalscatter}
        \mathbf{S}_{t} = \mathbf{XL}_{t}\mathbf{X}^{T} = \mathbf{X}(\mathbf{I} - \frac{1}{N}\mathbf{11}^{T})\mathbf{X}^{T},
    \end{equation}
    where $\mathbf{1}$ is a vector of ones.
    
    \item In third variant, denoted by GESSVDD-kNN, the $\mathbf{L}_{x}$ is replaced with the kNN graph $\mathbf{L}_{kNN}$, where \( \mathbf{L}_{kNN} \\ = \mathbf{D}_{kNN} - \mathbf{A}_{kNN} \). In this variant, we use the diagonal and adjacency matrices, denoted by $\mathbf{D}_{kNN}$ and $\mathbf{A}_{kNN}$, respectively. The elements of the $\mathbf{A}_{kNN}$ matrix are set to 1 if data points $\mathbf{x}_{i}$ or $\mathbf{x}_{j}$ are in each other's neighborhood, and 0 otherwise, mathematically expressed as
    
    \begin{equation}
    [\mathbf{A}_{ij}] = 
    \left\{
        \begin{array}{lr}
            1, \quad if \quad \mathbf{x}_{i} \in \mathbf{N}_{j} \quad or \quad \mathbf{x}_{j} \in \mathbf{N}_{i}\\
             0, \quad \text{otherwise}
        \end{array}
    \right\},
    \end{equation}
    where $\mathbf{N}_{i}$ represents the neighborhood of the $i^{th}$ data point.
\end{itemize}

Furthermore, all these variants have been solved using three different techniques: gradient-based, spectral, and spectral regression technique. In a gradient-based solution, given by $\mathbf{Q} \leftarrow \mathbf{Q} - \eta\Delta L$, the gradient of (\ref{gessvdd_lag}) is used to update the $\mathbf{Q}$. The variable $\eta$ is a hyper-parameter defining the step of the gradient. In contrast, the other two use eigenvalue and eigenvectors to find the optimized set of features \cite{sohrab2023graph}. The variants based on the solution technique are referred to by `G' for gradient-based, `E' for spectral, and `S' for spectral regression. Since there exists no rule of thumb to either maximize or minimize each solution, we experimented with both strategies (denoted by max and min, respectively) for SSVDD and GESSVDD models. In the gradient-based method, the ascending and descending steps in the update rule are used for maximizing and minimizing, respectively. In contrast, for the other two methods, the highest and lowest set of positive eigenvalues and corresponding eigenvectors are chosen for maximization and minimization, respectively. Moreover, different variants of the SSVDD model are implemented based on the regularization term $\Psi$ \cite{sohrab2018subspace}. A hyper-parameter $\beta$, which gives weight to the regularization term $\Psi$ in SSVDD, is tuned during cross-validation. Also, the non-linear version of all these models and variants is implemented using the non-linear projection trick (NPT) \cite{kwak2013nonlinear}. The kernel function utilized in the NPT is the Radial Basis Function (RBF), given by
\begin{equation}
    \label{kernel_func}
    \mathbf{K}_{ij} = exp \left ( -  \frac{\left\| \mathbf{x}_{i} - \mathbf{x}_{j}\right\|_{2}^{2}}{2\sigma^{2}}  \right ),
\end{equation}
where $\sigma$ is a hyper-parameter that defines the width of the kernel.

\section{Experiments and Discussion} \label{sec:experiments}

\begin{table*}[h]
\footnotesize\setlength{\tabcolsep}{6.5pt}
\begin{center}
\caption{Results for the linear versions of all models for all datasets. Pre stands for precision, F1 denotes F1-measure, G-m represents G-mean, and Avg of G-means is the average of G-means across the datasets. The highest performer in terms of G-mean for each dataset is marked in bold. The model names follow the following rule: for graph-based: [model]-[graph]-[solution method]-[min/max] and for SSVDD: [model]-[regularization term]-[min/max].}
\begin{tabular}{|l|llllllllllll|l|}
\hline
\multicolumn{1}{|c|}{\multirow{2}{*}{\textbf{Model}}} & \multicolumn{3}{c|}{\textbf{Dataset-1}}                                       & \multicolumn{3}{c|}{\textbf{Dataset-2}}                                       & \multicolumn{3}{c|}{\textbf{Dataset-3}}                                       & \multicolumn{3}{c|}{\textbf{Dataset-4}}                             & \multirow{2}{*}{\textbf{\begin{tabular}[c]{@{}l@{}}Avg of \\G-means\end{tabular}}} \\ \cline{2-13}
\multicolumn{1}{|c|}{}                                & \multicolumn{1}{c|}{Pre} & \multicolumn{1}{c|}{F1} & \multicolumn{1}{c|}{G-m} & \multicolumn{1}{c|}{Pre} & \multicolumn{1}{c|}{F1} & \multicolumn{1}{c|}{G-m} & \multicolumn{1}{c|}{Pre} & \multicolumn{1}{c|}{F1} & \multicolumn{1}{c|}{G-m} & \multicolumn{1}{c|}{Pre} & \multicolumn{1}{c|}{F1} & G-m            &                                                                                          \\ \hline
GESSVDD-kNN-G-min                                     & 1.000                    & 0.922                   & \textbf{0.906}           & 0.961                    & 0.523                   & 0.551                    & 0.999                    & 0.998                   & 0.603                    & 0.866                    & 0.664                   & 0.644          & \textbf{0.676}                                                                         \\
GESSVDD-kNN-G-max                                     & 1.000                    & 0.994                   & 0.849                    & 0.925                    & 0.766                   & 0.541                    & 0.999                    & 0.998                   & 0.603                    & 0.850                    & 0.803                   & \textbf{0.691} & 0.671                                                                                  \\
GESSVDD-kNN-E-min                                     & 0.999                    & 0.996                   & 0.640                    & 0.819                    & 0.373                   & 0.326                    & 1.000                    & 0.838                   & 0.697                    & 0.798                    & 0.743                   & 0.598          & 0.565                                                                                  \\
GESSVDD-kNN-E-max                                     & 0.999                    & 0.997                   & 0.686                    & 0.953                    & 0.896                   & \textbf{0.692}           & 0.999                    & 0.998                   & 0.603                    & 0.744                    & 0.687                   & 0.501          & 0.620                                                                                  \\
GESSVDD-kNN-S-min                                     & 0.998                    & 0.999                   & 0.296                    & 0.906                    & 0.675                   & 0.472                    & 1.000                    & 0.707                   & \textbf{0.728}           & 0.690                    & 0.623                   & 0.410          & 0.477                                                                                  \\
GESSVDD-kNN-S-max                                     & 0.998                    & 0.999                   & 0.296                    & 0.949                    & 0.089                   & 0.214                    & 1.000                    & 0.591                   & 0.638                    & 0.752                    & 0.769                   & 0.472          & 0.405                                                                                  \\
GESSVDD-PCA-G-min                                     & 1.000                    & 0.326                   & 0.432                    & 0.919                    & 0.955                   & 0.282                    & 0.999                    & 0.998                   & 0.595                    & 0.734                    & 0.844                   & 0.074          & 0.346                                                                                  \\
GESSVDD-PCA-G-max                                     & 1.000                    & 0.619                   & 0.644                    & 0.915                    & 0.953                   & 0.192                    & 0.999                    & 0.998                   & 0.605                    & 0.734                    & 0.844                   & 0.070          & 0.378                                                                                  \\
GESSVDD-PCA-E-min                                     & 0.999                    & 0.996                   & 0.777                    & 0.915                    & 0.953                   & 0.177                    & 0.999                    & 0.998                   & 0.624                    & 0.734                    & 0.845                   & 0.055          & 0.408                                                                                  \\
GESSVDD-PCA-E-max                                     & 0.999                    & 0.998                   & 0.720                    & 0.915                    & 0.954                   & 0.186                    & 0.999                    & 0.998                   & 0.595                    & 0.735                    & 0.845                   & 0.082          & 0.396                                                                                  \\
GESSVDD-PCA-S-min                                     & 0.998                    & 0.999                   & 0.164                    & 0.915                    & 0.954                   & 0.178                    & 0.999           & 0.998                   & 0.595                    & 0.745                    & 0.852                   & 0.241          & 0.294                                                                                  \\
GESSVDD-PCA-S-max                                     & 0.998                    & 0.999                   & 0.082                    & 0.915                    & 0.954                   & 0.193                    & 0.999           & 0.998                   & 0.595                    & 0.744           & 0.848                   & 0.243          & 0.279                                                                                  \\
GESSVDD-I-G-min                                       & 0.998                    & 0.998                   & 0.116                    & 0.901                    & 0.861                   & 0.209                    & 0.999                    & 0.998                   & 0.595                    & 0.718                    & 0.716                   & 0.401          & 0.330                                                                                  \\
GESSVDD-I-G-max                                       & 0.998                    & 0.999                   & 0.000                    & 0.914                    & 0.953                   & 0.123                    & 0.999                    & 0.998          & 0.595                    & 0.765                    & 0.748                   & 0.527          & 0.311                                                                                  \\
GESSVDD-I-E-min                                       & 0.999                    & 0.998                   & 0.725                    & 0.914                    & 0.953                   & 0.170                    & 0.999                    & 0.998                   & 0.624                    & 0.748                    & 0.787                   & 0.434          & 0.488                                                                                  \\
GESSVDD-I-E-max                                       & 0.999                    & 0.999                   & 0.493                    & 0.915                    & 0.954                   & 0.186                    & 0.999                    & 0.998                   & 0.595                    & 0.711                    & 0.668                   & 0.429          & 0.426                                                                                  \\
GESSVDD-I-S-min                                       & 0.998                    & 0.999                   & 0.164                    & 0.913                    & 0.951                   & 0.086                    & 0.999                    & 0.998                   & 0.595                    & 0.734                    & 0.844                   & 0.085          & 0.233                                                                                  \\
GESSVDD-I-S-max                                       & 0.998                    & 0.999                   & 0.082                    & 0.916                    & 0.954                   & 0.205                    & 0.999                    & 0.998                   & 0.610                    & 0.735                    & 0.843                   & 0.113          & 0.252                                                                                  \\
SSVDD-$\Psi_{0}$-min                                  & 0.999                    & 0.354                   & 0.438                    & 0.916                    & 0.954                   & 0.204                    & 0.999                    & 0.998                   & 0.407                    & 0.738                    & 0.847                   & 0.150          & 0.300                                                                                  \\
SSVDD-$\Psi_{0}$-max                                  & 0.999                    & 0.291                   & 0.391                    & 0.914                    & 0.954                   & 0.162                    & 0.999                    & 0.998                   & 0.404                    & 0.736                    & 0.846                   & 0.123          & 0.270                                                                                  \\
SSVDD-$\Psi_{1}$-min                                  & 0.998                    & 0.999                   & 0.000                    & 0.916                    & 0.954                   & 0.204                    & 0.999                    & 0.998                   & 0.407                    & 0.738                    & 0.847                   & 0.150          & 0.190                                                                                  \\
SSVDD-$\Psi_{1}$-max                                  & 0.998                    & 0.999                   & 0.000                    & 0.914                    & 0.954                   & 0.162                    & 0.999                    & 0.998                   & 0.404                    & 0.736                    & 0.846                   & 0.123          & 0.172                                                                                  \\
SSVDD-$\Psi_{2}$-min                                  & 0.989                    & 0.092                   & 0.183                    & 0.916                    & 0.954                   & 0.204                    & 0.999                    & 0.998                   & 0.407                    & 0.738                    & 0.847                   & 0.150          & 0.236                                                                                  \\
SSVDD-$\Psi_{2}$-max                                  & 0.989                    & 0.092                   & 0.183                    & 0.914                    & 0.954                   & 0.162                    & 0.999                    & 0.998                   & 0.404                    & 0.736                    & 0.846                   & 0.123          & 0.218                                                                                  \\
SSVDD-$\Psi_{3}$-min                                  & 0.989                    & 0.092                   & 0.182                    & 0.916                    & 0.954                   & 0.204                    & 0.999                    & 0.998                   & 0.407                    & 0.738                    & 0.847                   & 0.150          & 0.236                                                                                  \\
SSVDD-$\Psi_{3}$-max                                  & 0.989                    & 0.092                   & 0.182                    & 0.914                    & 0.954                   & 0.162                    & 0.999                    & 0.998                   & 0.404                    & 0.736                    & 0.846                   & 0.123          & 0.218                                                                                  \\
OCSVM                                                 & 0.999                    & 0.958                   & 0.446                    & 0.941                    & 0.599                   & 0.559                    & 1.000                    & 0.098                   & 0.227           & 0.582                    & 0.356                   & 0.355          & 0.397                                                                                  \\
SVDD                                                  & 0.993                    & 0.092                   & 0.198                    & 0.915                    & 0.954                   & 0.185                    & 0.999                    & 0.998                   & 0.404                    & 0.742                    & 0.848                   & 0.216          & 0.251                                                                                  \\
ESVDD                                                 & 0.000                    & 0.000                   & 0.000                    & 0.915                    & 0.954                   & 0.187                    & 0.999                    & 0.998                   & 0.595                    & 0.734                    & 0.847                   & 0.035          & 0.204                                                                                  \\ \hline
\end{tabular}
\label{table:results-1}
\end{center}
\end{table*}

\begin{table*}[h]
\footnotesize\setlength{\tabcolsep}{6.5pt}
\begin{center}
\caption{Results for the non-linear versions of all models for all datasets. Pre stands for precision, F1 denotes F1-measure, G-m represents G-mean, and Avg of G-means is the average of G-means across the datasets. The highest performer in terms of G-mean for each dataset is marked in bold. The model names follow the following rule: for graph-based: [model]-[graph]-[solution method]-[min/max] and for SSVDD: [model]-[regularization term]-[min/max].}
\begin{tabular}{|l|llllllllllll|l|}
\hline
\multicolumn{1}{|c|}{\multirow{2}{*}{\textbf{Model}}} & \multicolumn{3}{c|}{\textbf{Dataset-1}}                                       & \multicolumn{3}{c|}{\textbf{Dataset-2}}                                       & \multicolumn{3}{c|}{\textbf{Dataset-3}}                                       & \multicolumn{3}{c|}{\textbf{Dataset-4}}                             & \multirow{2}{*}{\textbf{\begin{tabular}[c]{@{}l@{}}Avg of \\ G-means\end{tabular}}} \\ \cline{2-13}
\multicolumn{1}{|c|}{}                                & \multicolumn{1}{c|}{Pre} & \multicolumn{1}{c|}{F1} & \multicolumn{1}{c|}{G-m} & \multicolumn{1}{c|}{Pre} & \multicolumn{1}{c|}{F1} & \multicolumn{1}{c|}{G-m} & \multicolumn{1}{c|}{Pre} & \multicolumn{1}{c|}{F1} & \multicolumn{1}{c|}{G-m} & \multicolumn{1}{c|}{Pre} & \multicolumn{1}{c|}{F1} & G-m            &                                                                                        \\ \hline
GESSVDD-kNN-G-min                                     & 0.998                    & 0.999                   & 0.329                    & 0.918                    & 0.955                   & 0.264                    & 0.999                    & 0.705                   & 0.570                    & 0.478                    & 0.278                   & 0.283          & 0.362                                                                                  \\
GESSVDD-kNN-G-max                                     & 0.970                    & 0.011                   & 0.070                    & 0.910                    & 0.199                   & 0.314                    & 0.999                    & 0.767                   & 0.559                    & 0.592                    & 0.435                   & 0.345          & 0.322                                                                                  \\
GESSVDD-kNN-E-min                                     & 0.294                    & 0.000                   & 0.007                    & 0.846                    & 0.304                   & 0.346                    & 0.999                    & 0.919                   & 0.576                    & 0.555                    & 0.381                   & 0.321          & 0.313                                                                                  \\
GESSVDD-kNN-E-max                                     & 0.000                    & 0.000                   & 0.000                    & 0.846                    & 0.304                   & 0.346                    & 0.999                    & 0.919                   & 0.576                    & 0.555                    & 0.381                   & 0.321          & 0.311                                                                                  \\
GESSVDD-kNN-S-min                                     & 0.999                    & 0.992                   & 0.522                    & 0.849                    & 0.441                   & 0.365                    & 0.999                    & 0.913                   & 0.580                    & 0.800                    & 0.005                   & 0.052          & 0.380                                                                                  \\
GESSVDD-kNN-S-max                                     & 0.999                    & 0.975                   & \textbf{0.550}           & 0.849                    & 0.441                   & 0.365                    & 0.999                    & 0.913                   & 0.580                    & 0.571                    & 0.009                   & 0.066          & 0.390                                                                                  \\
GESSVDD-PCA-G-min                                     & 1.000                    & 0.000                   & 0.003                    & 0.913                    & 0.954                   & 0.110                    & 0.999                    & 0.945                   & 0.592                    & 0.726                    & 0.770                   & 0.347          & 0.263                                                                                  \\
GESSVDD-PCA-G-max                                     & 0.999                    & 0.975                   & 0.446                    & 0.542                    & 0.001                   & 0.019                    & 0.999                    & 0.945                   & 0.592                    & 0.575                    & 0.407                   & 0.335          & 0.348                                                                                  \\
GESSVDD-PCA-E-min                                     & 0.988                    & 0.092                   & 0.181                    & 0.978                    & 0.552                   & 0.591                    & 0.999                    & 0.970                   & 0.547                    & 0.632                    & 0.026                   & 0.114          & 0.358                                                                                  \\
GESSVDD-PCA-E-max                                     & 0.998                    & 0.999                   & 0.000                    & 0.795                    & 0.338                   & 0.301                    & 0.999                    & 0.972                   & 0.533                    & 0.634                    & 0.026                   & 0.113          & 0.237                                                                                  \\
GESSVDD-PCA-S-min                                     & 0.999                    & 0.975                   & 0.417                    & 0.966                    & 0.640                   & 0.627                    & 0.996                    & 0.129                   & 0.227                    & 0.332                    & 0.031                   & 0.121          & 0.348                                                                                  \\
GESSVDD-PCA-S-max                                     & 0.998                    & 0.976                   & 0.341                    & 0.966                    & 0.640                   & 0.627                    & 0.999                    & 0.958                   & 0.579                    & 0.332                    & 0.031                   & 0.121          & 0.417                                                                                  \\
GESSVDD-I-G-min                                       & 0.250                    & 0.000                   & 0.003                    & 0.474                    & 0.008                   & 0.060                    & 0.999                    & 0.953                   & 0.555                    & 0.803                    & 0.072                   & 0.192          & 0.203                                                                                  \\
GESSVDD-I-G-max                                       & 0.999                    & 0.975                   & 0.494                    & 0.538                    & 0.004                   & 0.043                    & 0.999                    & 0.980                   & 0.601                    & 0.547                    & 0.356                   & 0.323          & 0.365                                                                                  \\
GESSVDD-I-E-min                                       & 0.222                    & 0.000                   & 0.007                    & 0.849                    & 0.346                   & 0.360                    & 0.999                    & 0.969                   & 0.557                    & 0.553                    & 0.009                   & 0.068          & 0.248                                                                                  \\
GESSVDD-I-E-max                                       & 0.998                    & 0.999                   & 0.000                    & 0.849                    & 0.346                   & 0.360                    & 0.999                    & 0.975                   & 0.497                    & 0.447                    & 0.036                   & 0.133          & 0.247                                                                                  \\
GESSVDD-I-S-min                                       & 0.998                    & 0.046                   & 0.151                    & 0.879                    & 0.149                   & 0.268                    & 0.999                    & 0.969                   & 0.557                    & 0.713                    & 0.599                   & 0.469          & 0.361                                                                                  \\
GESSVDD-I-S-max                                       & 0.998                    & 0.999                   & 0.000                    & 0.879                    & 0.149                   & 0.268                    & 0.999                    & 0.969                   & 0.557                    & 0.358                    & 0.035                   & 0.129          & 0.238                                                                                  \\
SSVDD-$\Psi_{0}$-min                                  & 0.999                    & 0.091                   & 0.215                    & 0.951                    & 0.277                   & 0.384                    & 1.000                    & 0.016                   & 0.089                    & 0.723                    & 0.733                   & 0.398          & 0.272                                                                                  \\
SSVDD-$\Psi_{0}$-max                                  & 0.996                    & 0.092                   & 0.208                    & 0.890                    & 0.315                   & 0.380                    & 0.996                    & 0.171                   & 0.244                    & 0.723                    & 0.733                   & 0.398          & 0.308                                                                                  \\
SSVDD-$\Psi_{1}$-min                                  & 0.992                    & 0.092                   & 0.194                    & 0.951                    & 0.277                   & 0.384                    & 1.000                    & 0.016                   & 0.089                    & 0.723                    & 0.733                   & 0.398          & 0.266                                                                                  \\
SSVDD-$\Psi_{1}$-max                                  & 0.999                    & 0.059                   & 0.173                    & 0.890                    & 0.315                   & 0.380                    & 0.996                    & 0.171                   & 0.244                    & 0.723                    & 0.733                   & 0.398          & 0.299                                                                                  \\
SSVDD-$\Psi_{2}$-min                                  & 0.992                    & 0.092                   & 0.193                    & 0.951                    & 0.277                   & 0.384                    & 1.000                    & 0.016                   & 0.089                    & 0.723                    & 0.733                   & 0.398          & 0.266                                                                                  \\
SSVDD-$\Psi_{2}$-max                                  & 0.999                    & 0.999                   & 0.386                    & 0.890                    & 0.315                   & 0.380                    & 0.996                    & 0.171                   & 0.244                    & 0.723                    & 0.733                   & 0.398          & 0.352                                                                                  \\
SSVDD-$\Psi_{3}$-min                                  & 0.993                    & 0.092                   & 0.197                    & 0.951                    & 0.277                   & 0.384                    & 1.000                    & 0.016                   & 0.089                    & 0.723                    & 0.733                   & 0.398          & 0.267                                                                                  \\
SSVDD-$\Psi_{3}$-max                                  & 0.996                    & 0.092                   & 0.208                    & 0.890                    & 0.315                   & 0.380                    & 0.996                    & 0.171                   & 0.244                    & 0.723                    & 0.733                   & 0.398          & 0.308                                                                                  \\
OCSVM                                                 & 1.000                    & 0.034                   & 0.131                    & 0.955                    & 0.577                   & 0.574                    & 1.000                    & 0.021                   & 0.102                    & 0.830                    & 0.822                   & 0.662          & 0.367                                                                                  \\
SVDD                                                  & 0.993                    & 0.092                   & 0.198                    & 0.214                    & 0.014                   & 0.073                    & 0.934                    & 0.003                   & 0.039                    & 0.284                    & 0.083                   & 0.179          & 0.122                                                                                  \\
ESVDD                                                 & 0.999                    & 0.089                   & 0.214                    & 0.348                    & 0.005                   & 0.048                    & 0.999                    & 0.945                   & 0.592                    & 0.185                    & 0.026                   & 0.108          & 0.241                                                                                  \\  
GEOCSVM                                                 & 0.999                    & 0.066                   & 0.183                    & 1.000                    & 0.936                   & \textbf{0.937}           & 1.000                    & 0.926                   & \textbf{0.791}           & 0.859                    & 0.828                   & \textbf{0.714} & \textbf{0.656}                                                                         \\
GESVDD                                                & 1.000                    & 0.089                   & 0.215                    & 0.997                    & 0.923                   & 0.913                    & 0.999                    & 0.882                   & 0.593                    & 0.849                    & 0.817                   & 0.694          & 0.604  \\ \hline
\end{tabular}
\label{table:results-2}
\end{center}
\end{table*}

\subsection{Datasets}
\label{sec:dataset}
In this paper, four datasets, all sourced from Kaggle\footnote{https://www.kaggle.com/datasets} open source dataset repository, are employed for evaluating the OCC models for detecting fraudulent credit card transactions. The first dataset, denoted by Dataset-1, originates from the Worldline and Machine Learning Group at Université Libre de Bruxelles (ULB). It includes credit card transactions made by European cardholders over two days in September 2013. It consists of 29 features and 284,807 transactions with only 492 fraudulent ones (which makes up 0.172\% of the dataset). The Dataset-2 contains digital payment transactions with 7 features and $1 \times 10^{6}$ instances, of which 87,403 are fraudulent. The imbalance ratio for this dataset is 0.087. The Dataset-3 is synthetically generated using the Paysim simulator based on a sample of real mobile transactions for one month. It comprises 5 features and 1,048,575 transactions, among which only 1142 are fraudulent, resulting in an imbalance ratio of 0.001. Lastly, a dataset from a bank, available at Kaggle, is utilized, which is denoted by Dataset-4. It includes 112 features and 20,467 transactions, with 5437 being fraudulent, representing 26.6\% of the dataset.

\subsection{Experimental Setup}
All datasets used in this study are split into 70-30 train-test sets. To handle the high number of instances in the training data, random resampling is performed while maintaining the skewed nature of the data. The resampled training Dataset-1 consists of 344 fraudulent and 2800 normal transactions, leading to a fraudulent-to-normal ratio of 0.12, whereas Dataset-2 to -4 consists of 500 fraudulent and 2500 normal 
transactions, giving a fraudulent-to-normal ratio of 0.2. Mean and standard deviation are calculated from target class data of the respective original (before resampling) dataset, which is used to normalize the reduced training dataset.

Model training involves tuning hyperparameters using 5-fold cross-validation over the training set. Performance metrics calculated and observed for this study are precision, F1-measure, and geometric mean of sensitivity and specificity (denoted by G-mean), but because of its balanced assessment of positive and negative instances, G-mean is used as an assessment metric during the cross-validation and for model evaluation. The iterative methods' number of iterations and the number of neighbors for the kNN graph are both set to 5. The hyperparameters tuned during cross-validation are given below:
\begin{itemize}
    \item $C \rightarrow$ [0.1 0.2 0.3 0.4 0.5]
    \item \textit{d} $\rightarrow$ [1 2 3 4 5 10 20]
    \item $\beta \rightarrow$ [0.01, 0.1, 1, 10, 100] 
    \item $\eta \rightarrow$ [0.1, 1, 10, 100, 1000] 
    \item $\sigma \rightarrow$ [0.1, 1, 10, 100, 1000]
\end{itemize}

\subsection{Results and Discussion}
The results for the linear and non-linear versions of the models for all datasets are given in Tables \ref{table:results-1} and \ref{table:results-2}, respectively. From the analysis of these results, it is evident that, for Dataset-1, the approach GESSVDD stands out. Particularly, its linear version and the utilization of the minimization-update rule exhibit notably better performance. The kNN graph and the gradient-based solution technique also outperform their counterparts for this specific dataset. For the other datasets, a non-linear model Graph-embedded One-Class Support Vector Machine GEOCSVM \cite{MYGDALIS2016585} displays better performance compared to other models. However, some models in each dataset exhibit a significantly high or low precision value. These models are either biased towards the positive class (in case of high values), or the boundary formed by these models is very small, and consequently, the normal transactions are forced out of the boundary and classified as fraudulent (in case of low values).

The analysis of the variants of SSVDD with regard to the regularization term, $\Psi$ shows that $\Psi_{0}$ produces more favorable results for Dataset-1. For the remaining datasets, all variants yield similar performance. Additionally, an overall assessment based on the average G-mean highlights the supremacy of $\Psi_{0}$, which indicates that, for the given datasets, incorporating the regularization term does not provide significant additional insights, and solving the conventional Lagrange equation suffices for optimization.

The analysis of graph-based vs. non-graph-based models shows that the integration of geometric information from the data yields enhanced performance. Consequently, models that leverage graph embeddings outperform those without such added information. Particularly, the kNN graph consistently outperforms other graph options considered for this study. Both eigenvalue decomposition and gradient-based solutions exhibit consistent performance across all datasets.

An investigation based on the average of G-means across datasets is performed to find the best-performing model and other strategies for all four datasets. It is found that GESSVDD with kNN graph, gradient-based solution, and minimization strategy in linear case works well on average for all datasets. It is also found that, on average, the linear version of the models outperforms the counter (non-linear) version. In contrast, the minimization or maximization update rule does not have a significant effect on the performance of the model. Moreover, it is also established that, in general, the kNN graph works better than the other graphs considered in the study.

\section{Conclusion}
Detecting credit card fraud remains a challenge despite the advances in technology. The imbalanced data and evolving fraud techniques contribute to this difficulty. Additionally, the curse of dimensionality is a challenge that poses problems for feature extraction. To address these issues without altering data proportions synthetically, we employed OCC algorithms, particularly subspace learning-based models. These models efficiently learn patterns in the data and predict fraudulent transactions, reducing losses caused by fraud.

In this research, we used four imbalanced datasets from Kaggle, resampled while retaining the data's imbalanced nature. We trained 60 model variants, including OCSVM, GEOCSVM, SVDD, GESVDD \cite{MYGDALIS2016585}, ESVDD, SSVDD, and GESSVDD. Results show that, on average, the linear GESSVDD with kNN graph, gradient-based solution, and minimization-update rule outperforms other models for all datasets. The G-mean metric is used for model evaluation based on its balanced assessment of both positive and negative instances.

Due to the high complexity of the models, high computational power is required to train the models, calling for improved complexity and efficiency. Additionally, the lack of real-world datasets due to data privacy rules hinders the interpretability and extraction of meaningful features by handcrafted methods. Future work involves investigating other kernel types and graphs for existing methods for improved results. In the future, we plan to adapt Multi-modal Subspace Support Vector Data Description (MSSVDD) \cite{sohrab2021multimodal} for credit card fraud detection.

\section*{ACKNOWLEDGEMENT}
This work was supported by the NSF-Business Finland project AMALIA. Foundation for Economic Education (Grant number: 220363) funded the work of Fahad Sohrab at Haltian.
\bibliography{refs}
\bibliographystyle{ieeetr}
\end{document}